\documentclass[submission]{eptcs}
\providecommand{\event}{DC, ICLP 2019} 
\usepackage{breakurl}             
\usepackage{graphicx}
\usepackage{algorithm,algpseudocode}
\usepackage{adjustbox}
\newtheorem{exmp}{Example}
\newtheorem{definition}{Definition}
\algnewcommand\algorithmicinput{\textbf{Input:}}
\algnewcommand\Input{\item[\algorithmicinput]}
\algnewcommand\algorithmicoutput{\textbf{Output:}}
\algnewcommand\Output{\item[\algorithmicoutput]}
\algnewcommand\algorithmicforeach{\textbf{for each}}
\algdef{S}[FOR]{ForEach}[1]{\algorithmicforeach\ #1\ \algorithmicdo}

\title{Induction of Non-monotonic Logic Programs To Explain Statistical Learning Models}
\author{Farhad Shakerin
\institute{Department of Computer Science}
\institute{The University of Texas at Dallas\\ Richardson, USA}
\email{farhad.shakerin@utdallas.edu}}
\def\titlerunning{Induction of NMLPs To Explain Statistical Learning Models}
\def\authorrunning{F. Shakerin}
\begin{document}
\maketitle

\section{Introduction \& Problem Description}
Dramatic success of machine learning (ML) has led to a torrent of Artificial Intelligence (AI) applications. However, the effectiveness of these systems is limited by the machines' current inability to explain their decisions and actions to human users. That's mainly because  statistical machine learning methods produce models that are complex algebraic solutions to optimization problems such as risk minimization
or data likelihood maximization. Lack of intuitive descriptions makes it hard for users to understand and verify the underlying rules that govern the model. Also,
these methods cannot produce a justification for a prediction
they compute for a new data sample. It is important to be able to ``explain" the model that has been learned by a given machine learning technique. That is, as ML researchers/practitioners, we want to find the logic---approximated by a set of rules, for example---that explains the behavior of the learned model to the user. This explanation can be very useful to users as they can not only develop an intuitive understanding of the model, it can lead to further actions. This intuitive understanding can allow a banker, for example, to advise a loan applicant what changes they need to make to their profile in order to move from ``loan rejected" classification to ``loan approved" one. Explanations are also being necessitated by regulatory requirements such as the European GDPR \cite{gdpr}. Explanations can also reveal bias in the learned model, as well as improve a users' trust in the machine learning system's output.

Significant effort is being invested towards ``Explainable AI" \cite{xai}. The goal of these endeavors is to create a suite of machine learning techniques that: a) Produce more explainable models, while maintaining a high level of prediction accuracy. b) Enable human users to understand, appropriately trust, and effectively manage the emerging generation of AI systems. In this project, we propose to conduct research to achieve these goals.

Inductive Logic Programming (ILP) \cite{ilp} is one Machine Learning technique where the learned model is in the form of logic programming rules (Horn Clauses) that are comprehensible to humans. It allows the background knowledge to be incrementally extended without requiring the entire model to be re-learned. Meanwhile, the comprehensibility of symbolic rules makes it easier for users to understand and verify induced models and even refine them.

However, due to lack of \textit{negation-as-failure} \cite{baral}, Horn clauses are not sufficiently expressive for representation and reasoning when the background knowledge is incomplete. Additionally, ILP is not able to handle exception to general rules: it
learns rules under the assumption that there are no exceptions to them. This results in exceptions and noise being treated in the same manner. The resulting theory that is learned is a \textit{default theory} \cite{gelfond-book,baral}, and in most cases this theory describes the underlying model more accurately. It should be
noted that default theories closely model common sense reasoning as well \cite{baral}. Thus, a default theory, if it can be learned, will be more intuitive and comprehensible for humans. Default reasoning also allows us to reason in absence of information. A system that can learn default theories can therefore learn rules that can draw
conclusions based on lack of evidence, just like humans.
Other reasons that underscore the importance of inductive learning of default theories can be found in Sakama \cite{sakama05} who also surveys other attempts in this direction.

In this dissertation, I am investigating heuristics-based algorithms to learn \textit{non-monotonic} logic programs, (i.e., logic programs (Horn clauses) extended with \textit{negation-as-failure} \cite{gelfond-book,baral}). My novel algorithm extends the famous FOIL algorithm \cite{foil} by Quinlan. The work resulted in an algorithm called FOLD (\textbf{F}irst \textbf{O}rder \textbf{L}earner of \textbf{D}efault-theories) that was published in \textit{Theory and practice of Logic Programming} \cite{fold}. Because of the presence of \textit{negation-as-failure}, the answer set programs that are learned by FOLD, are more succinct and concise, compared to standard logic programs without negation. They also capture the underlying logic more accurately.

\begin{figure}
\centering
    \includegraphics[width=0.7\textwidth,scale=0.2]{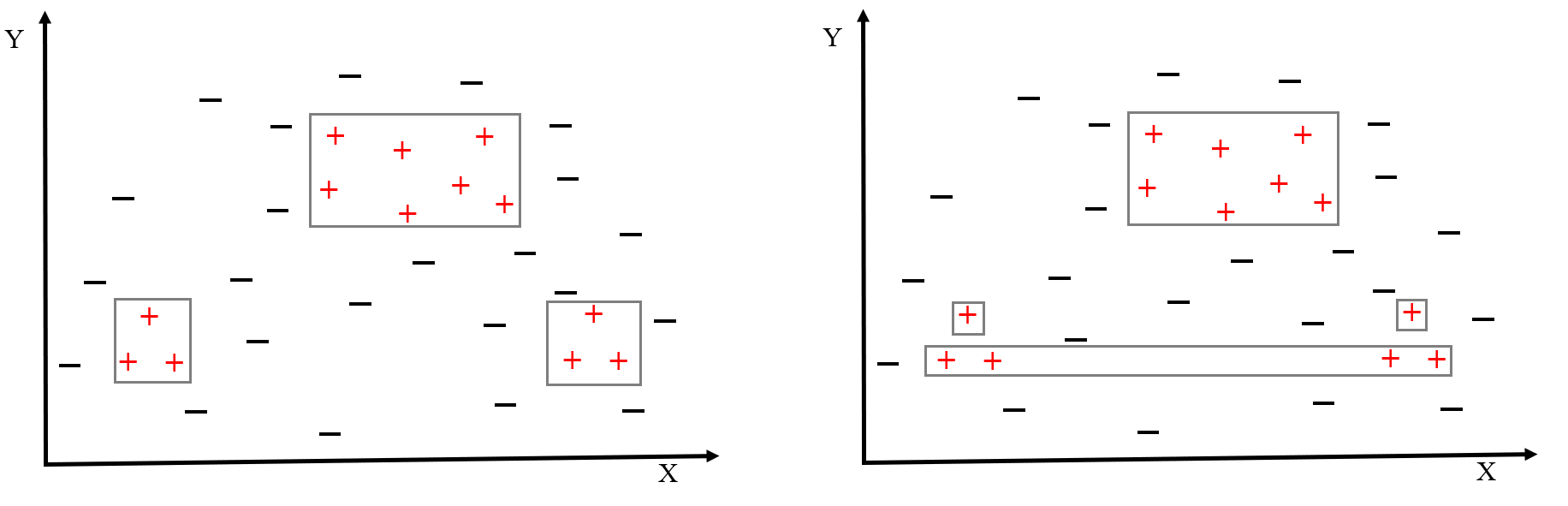}
\caption{Optimal sequential covering with 3 Clauses (Left), Sub-Optimal sequential covering with 4 Clauses (Right)}
\label{fig:localoptima}
\end{figure}

The FOIL algorithm itself is a popular top-down algorithm. FOIL uses  heuristics from information theory called \textit{weighted information gain}. The use of a greedy heuristic allows FOIL to run much faster than bottom-up approaches and scale up much better. However, scalability comes at the expense of losing accuracy if the algorithm is stuck in a local optima and/or when the number of examples is insufficient. Figure \ref{fig:localoptima} demonstrates how the local optima results in discovering sub-optimal rules that does not necessarily coincide with the real underlying sub-concepts of the data. Additionally, since the objective is to learn pure clauses (i.e., clauses with zero or few negative example coverage) the FOIL algorithm often discovers too many clauses each of which only cover a few examples. Discovering a huge number of clauses reduces the interpretability and also it does not generalize well on the test data.

Unlike top-down ILP algorithms, statistical machine learning algorithms are bound to find the relevant features because they optimize an objective function with respect to global constraints. This results in models that are inherently complex and cannot explain what features account for a classification decision on any given data sample. The Explainable AI techniques such as LIME \cite{lime} and SHAP \cite{shap} have been proposed that provide explanations for any given data sample. Each explanation is a set of feature-value pairs that would locally determine what features and how strongly each feature, relative to other features, contributes to the classification decision. To capture the global behavior of a black-box model, however, an algorithm needs to group similar data samples (i.e., data samples for which the same set of feature values are responsible for the choice of classification) and cover them with the same clause.

However, the FOLD algorithm although capable of learning more concise and accurate logic programs, still suffers from the local optima. Another major contribution of my dissertation is to create heuristics based on statistical learning models (as opposed to information gain and other data dependent heuristics) that would guide the search for the ``best" clause using insights taken from a statistical model such as an SVM model or a random forest. The resulted algorithms do not suffer the local optima. 

\section{Background}
\subsection{Inductive Logic Programming (ILP)}

Inductive Logic Programming (ILP) \cite{ilp} is one Machine Learning technique where the learned model is in the form of logic programming rules (Horn Clauses) that are comprehensible to humans. It allows the background knowledge to be incrementally extended without requiring the entire model to be re-learned. Meanwhile, the comprehensibility of symbolic rules makes it easier for users to understand and verify induced models and even edit them. The ILP problem is formally defined as follows:\\
\textbf{Given}
\begin{enumerate}
    \item a background theory $B$, in the form of an extended logic program, i.e., clauses of the form $h \leftarrow l_1, ... , l_m,\ not \ l_{m+1},...,\ not \ l_n$, where $l_1,...,l_n$ are positive literals and \textit{not} denotes \textit{negation-as-failure} (NAF) and $B$ has no even cycle
    \item two disjoint sets of ground target predicates $E^+, E^-$ known as positive and negative examples respectively
    \item a hypothesis language of function free predicates $L$, and a  refinement operator $\rho$ under $\theta-subsumption$ \cite{plotkin70} that would disallow even cycles.
\end{enumerate}
\textbf{Find} a set of clauses $H$ such that:
\begin{itemize}
    \item $ \forall e \in \ E^+ ,\  B \cup H \models e$
    \item $ \forall e \in \ E^- ,\  B \cup H \not \models e$
    \item $B$ and $H$ are consistent.
\end{itemize}

The ILP learning problem can be regarded as a search problem for a set of clauses from which training examples can be \textit{deduced}. The search is performed either top down or bottom-up. A bottom-up approach builds most-specific clauses from the training examples and searches the hypothesis space by using generalization. This approach is not applicable to large-scale datasets, nor it can incorporate \textit{Negation-As-Failure} into the hypotheses. A survey of bottom-up ILP systems and their shortcomings can be found at \cite{sakama05}. In contrast, top-down approach starts with the most general clauses and then specializes them. A top-down algorithm guided by heuristics is better suited for large-scale and/or noisy datasets \cite{quickfoil}, particularly, because it is scalable.

The FOIL algorithm by Quinlan \cite{foil} is a popular top-down algorithm. FOIL uses  heuristics from information theory called \textit{weighted information gain}. The use of a greedy heuristic allows FOIL to run much faster than bottom-up approaches and scale up much better. For instance, the QuickFOIL system \cite{quickfoil} can deal with millions of training examples in a reasonable time.

FOIL is a top-down ILP algorithm which follows a \textit{sequential covering} approach to induce a hypothesis. The FOIL algorithm is summarized in Algorithm \ref{algo:foil}. This algorithm repeatedly searches for clauses that score best with respect to a subset of 
positive and negative examples, a current hypothesis and a heuristic called \textit{information gain} (IG). 

\begin{algorithm}[ht]
\caption{Summarizing the FOIL algorithm}
\label{algo:foil}
\begin{algorithmic}[1]
\Input $target,B,E^+,E^-$ 
\Output H
\State $H \gets \emptyset $
\While{($|E^+| > 0$)}
	\State $c \gets (target$ :- $ \ true.)$
	\While{($|E^-| > 0 \land c.length < max\_length $)}
		\For{all $ \ c' \in \rho (c)$}
        	\State $compute \ score(E^+,E^-,H \cup \{c'\},B)$
    	\EndFor
    	\State let $\hat{c}$ be the $c' \in \rho(c)$ with the best score   
         \State $E^- \gets covers(\hat{c},E^-)$
    \EndWhile	
    \State add $\hat{c}$ to $H$
    \State $E^+ \gets E^+ \setminus covers(\hat{c},E^+)$
\EndWhile 
\State \textbf{return} $H$
\end{algorithmic}
\end{algorithm}

The inner loop searches for a clause with the highest information gain using a general-to-specific hill-climbing search. To specialize a given clause $c$, a refinement operator $\rho$ under $\theta$-subsumption ~\cite{plotkin70} is employed.
\begin{definition}
Clause C $\theta$-subsumes clause D, denoted by $C \preceq D$, if there exists a substitution $\theta$ such that $head(C)\theta = head(D)$ and $body(C)\theta \subseteq body(D)$, where $head(C)$ and $body(C)$ denote the head and the body of clause C respectively. $\theta$-subsumption provides a syntactic notion of generality that is used in \textit{general-to-specific} search for clauses based on their example coverage.
\end{definition}

The most general clause is $p(X_1,...,X_n) \gets true.$ where the predicate $p/n$ is the predicate being learned and each $X_i$ is a variable. The refinement operator based on $\theta$-subsumption specializes the current clause $h \gets b_1,...b_n .$ This is realized by adding a new literal $l$ to the clause yielding $h \gets b_1,...b_n,l$. The heuristic based search uses information gain. In FOIL, information gain for a given clause is calculated as follows: 
\begin{equation}
IG(L,R) = t\left(log_2 \frac{p_1}{p_1 + n_1} - log_2 \frac{p_0}{p_0+ n_0} 
\right)
\end{equation}
where $L$ is the candidate literal to add to rule $R$, $p_0$ is the number of 
positive bindings of $R$, $n_0$ is the number of negative bindings of $R$, 
$p_1$ is the number of positive bindings of $R+L$, $n_1$ is the number of 
negative bindings of $R+L$, $t$ is the number of possible bindings that make the
clause cover positive examples,

FOIL handles negated literals in a naive way by adding the literal $not \ L$ to 
the set of specialization candidate literals for any existing candidate $L$. 
This approach leads to learning predicates that do not capture the concept accurately as shown in the following example.

\begin{exmp}
$ \mathcal{B}, \mathcal{E^+}$ are background knowledge and positive examples 
respectively with CWA and the concept to be learned is fly.

\begin{tabular}{cll}
$\mathcal{B}: $    & $bird(X) \leftarrow penguin(X).$ & \\
                   & $bird(tweety).$                  & $bird(et).$ \\
                   & $cat(kitty).$                    & $penguin(polly).$\\
$\mathcal{E^+}:$   & $fly(tweety).$                   & $fly(et).$ \\

\end{tabular}
\end{exmp}
The FOIL algorithm would learn the following rule:
\[ fly(X) \leftarrow not \ cat(X), not \ penguin(X).\]
which does not yield a constructive definition even though it covers all the 
positives (tweety and et are not penguins and cats resp.) and no negatives 
(neither cats nor penguins do not fly). In fact, the correct theory in this 
example is  as follows: "{\it Only birds fly but, among them there are 
exceptional ones who do not fly}". It translates to the following logic programming rule:
\begin{verbatim}
fly(X) :- bird(X), not penguin(X).
\end{verbatim}
which FOIL fails to discover.

\subsection{The FOLD Algorithm}

We first present our FOLD algorithm that we have developed to learn default theories from background knowledge, positive and negative examples. The FOLD algorithm which is an extension of FOIL, learns a concept in terms of a default and possibly multiple exceptions (and exceptions to exceptions, exceptions to exceptions of exceptions, and so on). FOLD tries first to learn the default by specializing a general rule of the form \texttt{\{target($V_1,...,V_n$) :- true.\}} with positive literals. 
As in FOIL, each specialization must rule out some already covered negative 
examples without decreasing the number of positive examples covered 
significantly. Unlike FOIL, no negative literal is used at this stage. Once 
the heuristic score (i.e., \textit{information gain}) (IG) becomes zero, or the maximum clause length is reached (whichever happens first), this process stops. At this point, if any negative 
example is still covered, they must be either noisy data or 
exceptions to the current hypothesis. Exceptions could be learned by swapping the current positive and negative examples, then calling the same algorithm recursively. As a result of this recursive process, FOLD can learn exceptions to exceptions, and so on. In presence of noise, FOLD identifies and \textit{enumerates} noisy samples, that is, outputs them as ground facts in hypothesis, to make sure that the algorithm converges. \textit{Maximum Description Length Principle} \cite{foil} is incorporated to heuristically control the hypothesis length and identify noise. Algorithm 3 presents the pseudo-code of the FOLD algorithm.

{\small
\begin{algorithm}[!ht]
\caption{FOLD Algorithm}
\label{algo:fold}
\begin{algorithmic}[1]
\Input $target, B,E^+,E^-$ 
\Output  $D = \{c_1,...,c_n\}$   ~~~\%{defaults clauses}
\Statex $AB = \{ab_1,...,ab_m\}$ ~~~~~\%{exceptions clauses}
\Function{FOLD}{$E^+,E^-$} 
\While{($|E^+| > 0$)}
\State $c \gets (target$ :- $ \ true.)$
\State $\hat{c} \gets$ \Call{specialize}{{c},{$E^+$},{{$E^-$}}}
\State $E^+ \gets E^+ \setminus 
covers(\hat{c},E^+,B)$
\State $D \gets D \cup \{ \hat{c} \}$
\EndWhile 
\EndFunction
\Function{SPECIALIZE}{${c},{E^+},{E^-}$}
\While{$|E^-|>0 \land c.length < max\_rule\_length$}
\State  $(c_{def},\hat{IG}) \gets$ \Call{add\_best\_literal}{{c},{$E^+$},{{$E^-$}}}
\If{$\hat{IG} > 0$}
	\State $\hat{c} \gets c_{def} $
\Else
	\State $\hat{c} \gets \Call{exception}{{c},{E^-},{{E^+}}}$
			 \If {$\hat{c} == null$}
						\State $\hat{c} \gets enumerate(c,E^+)$
			\EndIf	
\EndIf
\State $E^+ \gets E^+ \setminus 
covers(\hat{c},E^+,B)$
\State $E^- \gets covers(\hat{c},E^-,B)$
\EndWhile
\EndFunction

\Function{EXCEPTION}{${c_{def}},{E^+},{E^-}$}
\State  $\hat{IG} \gets 
\Call{add\_best\_literal}{{c},{E^+},{{E^-}}}$
\If{$\hat{IG} > 0$}
	\State $ c\_set \gets \Call{FOLD}{E^+,E^-} $
	\State $ c\_ab \gets generate\_next\_ab\_predicate()$
	\ForEach {$c \in c\_set $}
		\State $AB \gets AB \cup \{ c\_ab $:-$ \ bodyof(c) \}$
	\EndFor
	\State $\hat{c} \gets (headof(c_{def}) $:-$ \ bodyof(c), 
\textbf{not}(c\_ab))$
\Else
	\State $\hat{c} \gets null$
\EndIf

\EndFunction
\end{algorithmic}
\end{algorithm}
}

\begin{exmp}
\label{ex:pinguin}
$ B, E^+$ are background knowledge and positive examples 
respectively under \textit{Closed World Assumption}, and the target predicate is \textit{fly}.
\end{exmp}
\begin{verbatim}
B:  bird(X) :- penguin(X).
    bird(tweety).   bird(et).
    cat(kitty).     penguin(polly).
E+: fly(tweety).    fly(et).
\end{verbatim}
Now, we illustrate how FOLD discovers the above set of clauses given 
$E^+ = \{tweety,et\}$ and $E^- = \{polly,kitty\}$ and the 
target \texttt{fly(X)}. By calling FOLD, at line 2 while loop, the clause 
\texttt{\{fly(X) :- true.\}} is specialized. Inside the $SPECIALIZE$ function, 
at line 10, the 
literal \texttt{bird(X)} is selected to add to the current clause, to get the 
clause 
$\hat{c}$ = \texttt{fly(X) :- bird(X)}, which happens to have the greatest IG 
among \texttt{\{bird,penguin,cat\}}. Then, at lines 20-21 the following updates 
are 
performed: $E^+=\{\}$,\ $E^-=\{polly\}$. A negative example 
$polly$, a penguin is still covered. In the next iteration, $SPECIALIZE$ fails 
to introduce a positive literal to rule it out since the best IG in this case 
is zero. Therefore, the EXCEPTION function is called by swapping the 
$E^+$, $E^-$. Now, FOLD is recursively called to learn a 
rule for $E^+ = \{polly\}$, $E^-=\{\}$. The recursive call 
(line 27), returns \texttt{\{fly(X) :- penguin(X)\}} as the exception. In line 
28, 
a new predicate \texttt{ab0} is introduced and at lines 29-31 the clause 
\texttt{\{ab0(X) :- penguin(X)\}} is created and added to the set of invented 
abnormalities, namely, AB. In line 32, the negated exception (i.e \texttt{not 
ab0(X)}) and the default rule's body (i.e \texttt{bird(X)}) are compiled 
together to form the following theory:
\begin{center}
    \begin{tabular}{l}
        \texttt{fly(X) :- bird(X), not ab0(X).}\\     
        \texttt{ab0(X) :- penguin(X).}     
    \end{tabular}
\end{center}

\subsection{The LIME Approach}
\label{sec:lime}
LIME \cite{lime} is a novel technique that finds easy to understand explanations for the predictions of any complex black-box classifier in a faithful manner. LIME constructs a linear model by sampling $N$ instances around any given data sample $x$. Every instance $x'$ represents a perturbed version of $x$ where perturbations are realized by sampling uniformly at random for each feature of $x$. LIME stores the classifier decision $f(x')$ and the kernel $\pi(x,x')$. The  $\pi$ function measures how similar the original and perturbed sample are and it is then used as the associated weight of $x'$ in fitting a \textit{locally weighted linear regression} (LWR) curve around $x$. The $K$ greatest learned weights of this linear model are interpreted as top $K$ contributing features into the decision made by the black-box classifier. Algorithm \ref{algo:lime} illustrates how a locally linear model is created around $x$ to explain a classifier's decision.

\begin{figure}[h]
    \includegraphics[width=\linewidth]{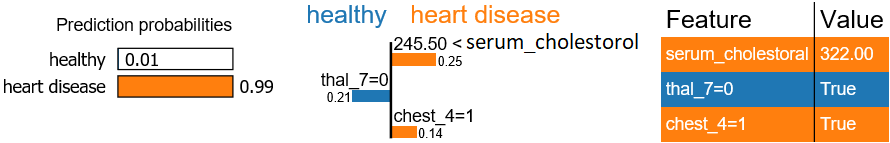}
\caption{Top 3 Relevant Features in Patient Diagnosis According to LIME}
\label{fig:heartlime}
\end{figure}

\begin{algorithm}[ht]
\caption{Linear Model Generation by LIME}
\label{algo:lime}
\begin{algorithmic}[1]
\Input $f:$ Classifier
\Input $N:$ Number of samples, $K:$ length of explanation,
\Input $x:$ sample to explain, $\pi :$ similarity kernel
\Output $w:$ Importance vector of features 
\State $\mathcal{Z} \gets \{\}$
\For{$ i \in \{1,2,3,...,N\}$}
	\State $x'_i \gets sample\_around(x)$
	\State $\mathcal{Z} \gets \mathcal{Z} \cup \langle x'_i,f(x_i),\pi(x_i) \rangle$ 
\EndFor
\State $w \gets LWR(\mathcal{Z},K)$
\State \textbf{return} $w$
\end{algorithmic}
\end{algorithm}
\noindent
The interpretation language should be understandable by humans. Therefore, LIME requires the user to provide some interpretation language as well. In case of tabular data, it boils down to specifying the valid range of each table column. In particular, if the data column is a numeric variable (as opposed to categorical), the user must specify the intervals or a discretization strategy to allow LIME to create intervals that are used later on to explain the classification decision.
\begin{exmp}
\label{ex:heart}
The UCI heart dataset contains features such as patient's  blood pressure, chest pain, thallium test results, number of major vessels blocked, etc. The classification task is to predict whether the subject suffers from heart disease or not. Figure \ref{fig:heartlime} shows how LIME would explain a model's prediction over a data sample.   
\end{exmp}
In this example, LIME is called to explain why the model predicts heart disease. In response, LIME returns the top features along with their importance weight. According to LIME, the model predicts ``heart disease" because of high serum cholesterol level, and having a chest pain of type 4 (i.e., asymptomatic). In this dataset, chest pain level is a categorical variable with 4 different values.

The categorical variables should be \textit{propositionalized} before a statistical model can be applied. Propositionalization is the process of transforming each categorical variable with domain of cardinality $n$, into $n$ new binary features. The feature ``thallium test'' is a categorical feature too. However, in this case LIME reports that the feature ``thal\_7" which is a new feature that resulted from propositionalization and has the value ``false'', would have made the model predict ``healthy". The value 7 for thallium test in this dataset indicates reversible defect which is a strong indication of heart disease. It should be noted that the feature ``serum cholesterol" is discretized with respect to the training examples' label. Discretization aims to reduce the number of values a continuous variable takes by grouping them into intervals or bins.

In order to capture model's global behavior, we are investigating an approach to learn concise logic programs from a transformed data set that is generated by storing the explanations provided by LIME. The LIME system takes 
as input a black-box model (such as a Neural Network, Random Forest, etc.) and a data sample. For any given data sample, it outputs a list of (weighted) features that contribute most to the classification decision. By repeating the same process for all training samples, we can generate a transformed version of the original data set that only contains the relevant features for each data sample.

\section{Research Goals}

The goal is to develop a heuristic-based ILP algorithm that would replace the traditional \textit{weighted information gain} heuristic with a new one. The new heuristic should guide the search based on the insights that a statistical learning model provides. We have extended the FOLD algorithm to incorporate the information generated by the pioneering LIME system \cite{lime}, which essentially computes the features that are deemed most important in classifying a given data sample. With help of LIME, our algorithm (called LIME-FOLD \cite{AAAI2018}) is able to learn answer set programs that are significantly more accurate and concise compared to standard ILP methods (including Aleph \cite{aleph} and our own FOLD method). For example, for the ``who makes 50K" UCI Adult data set, the number of learned rules go down from nearly 130 on the Aleph system \cite{aleph} to just 6 under LIME-FOLD while simultaneously improving precision, recall, accuracy and the F1-score. 

\section{Current Research Status}
I have further improved the application of explainable AI tools to capture the global behavior of statistical learning models by incorporating SHAP \cite{shap} and High Utility Item-set Mining (HUIM) \cite{huim} which is a powerful technique in datamining. The idea is to group similar data samples (i.e., data samples for which the same set of feature values are responsible for the choice of classification) and cover them with the same clause. While in FOIL, the search for a clause is guided by heuristics, in this novel approach, I adapt \textit{High Utility Item-set Mining} (HUIM) \cite{huim} --- a popular technique from data mining --- to find clauses. We call this algorithm SHAP-FOLD. The standard FOLD algorithm, LIME-FOLD and SHAP-FOLD have all been implemented and are available online at \cite{farhadgithub}

\section{Open Issues}

To make a Machine Learning model fully explainable, not only all the decisions need to be explainable, but also according to GDPR, an automated decision making system should also provide a user with minimum necessary efforts to flip a decision. This is known as \textit{Counter-factual Explanation}.  A counterfactual explanation describes a causal situation in the form: ``If X had not occurred, Y would not have occurred''. In interpretable machine learning, counterfactual explanations are minimal necessary changes in the feature values of an individual data sample to flip the prediction of a classifier from one class to another. In particular this is important in financial and educational applications where an automated system should provide recommendations as to how go from the state of ``rejection'' to ``approval" with minimum required efforts. The existing solutions based linear programming optimisation, do not consider the physical and logical constraints when they try to find solutions \cite{counterfactual}. The declarative nature of ILP-induced clauses have the potential to explain why the hypothesis holds for a particular data sample. One can go a step further than just explaining a machine learning model: one can provide advise on how to change the classification of a sample data. For example, if a machine learning based model is used to decide whether a loan application ought to be approved or not, one can compute with the help of Craig interpolant \cite{mcmillan} how various features must change in order for that application to become qualified. Significantly greater understanding of the model can thus be gained with the help of Craig interpolants. 

\nocite{*}
\bibliographystyle{eptcs}
\bibliography{generic}
\end{document}